\documentclass{article}
\usepackage{spconf,amsmath,amssymb,graphicx,hyperref,booktabs}

\title{Reasoning Instructions Can Break Answer Decoding in Vision--Language Models}
\name{Zeyan Li$^1$, Siyuan Qiu$^1$, Jianfeng Xu$^1$\sthanks{Corresponding author.}}
\address{$^1$ Shanghai Jiao Tong University}
\begin{document}
\ninept
\maketitle
\begin{abstract}
Chain-of-thought (CoT) instructions can distort multiple-choice VLM evaluation when a scorer appends a reasoning cue but reads answer-label logits before the model generates any rationale. We call this CoT-prefix scoring. On ScienceQA, Qwen2.5-VL-7B drops from 80.76\% to 45.48\%, and across five option-content permutations 93.54\% of CoT-prefix predictions select the first slot. Condition-matched linear probes recover 78.94\% from the same hidden states, while free generation restores 75.24\%, showing that the answer often survives the prefix and the immediate readout fails. Vocabulary and layer diagnostics explain the mismatch: probability mass moves toward continuation tokens, while answer information remains linearly accessible in late layers. The effect recurs with varying severity across datasets and models, though not universally. These results show that CoT-prefix scoring can confound model knowledge with an evaluation-interface mismatch and should be avoided unless the requested and scored output events are aligned.
\end{abstract}
\begin{keywords}
vision--language models, chain of thought, multiple-choice evaluation, selection bias
\end{keywords}
\section{Introduction}
\label{sec:intro}

Chain-of-thought (CoT) prompting lets a model generate intermediate steps before it answers, and this idea has been widely adopted in multimodal reasoning \cite{wei2022chain,lu2022learn,tian2025more}. In multiple-choice evaluation, however, a common shortcut skips the reasoning. The evaluator appends an instruction such as ``Let me think step by step.'' to the question, then reads the answer directly from the next-token logits on the option labels, without waiting for the model to generate any explanation. The prompt asks the model to start explaining, while the scorer treats its first token as the final answer. We refer to this procedure as \emph{CoT-prefix scoring}. It is a scoring convention that has been widely used in practice, but it has not been systematically examined.

We find that this shortcut can severely understate a strong model. On ScienceQA \cite{lu2022learn}, Qwen2.5-VL-7B \cite{bai2025qwen25vl} loses more than thirty points of accuracy under CoT-prefix scoring, and the failure is content-blind. Across five permutations of the option contents, the large majority of predictions land on the first option regardless of what that option contains. Nothing about the questions has changed, only the suffix. Either the suffix has erased the answer from the model, or the answer remains inside and the one-step readout cannot reach it.

\begin{figure}[t]
\centering
\includegraphics[width=\columnwidth]{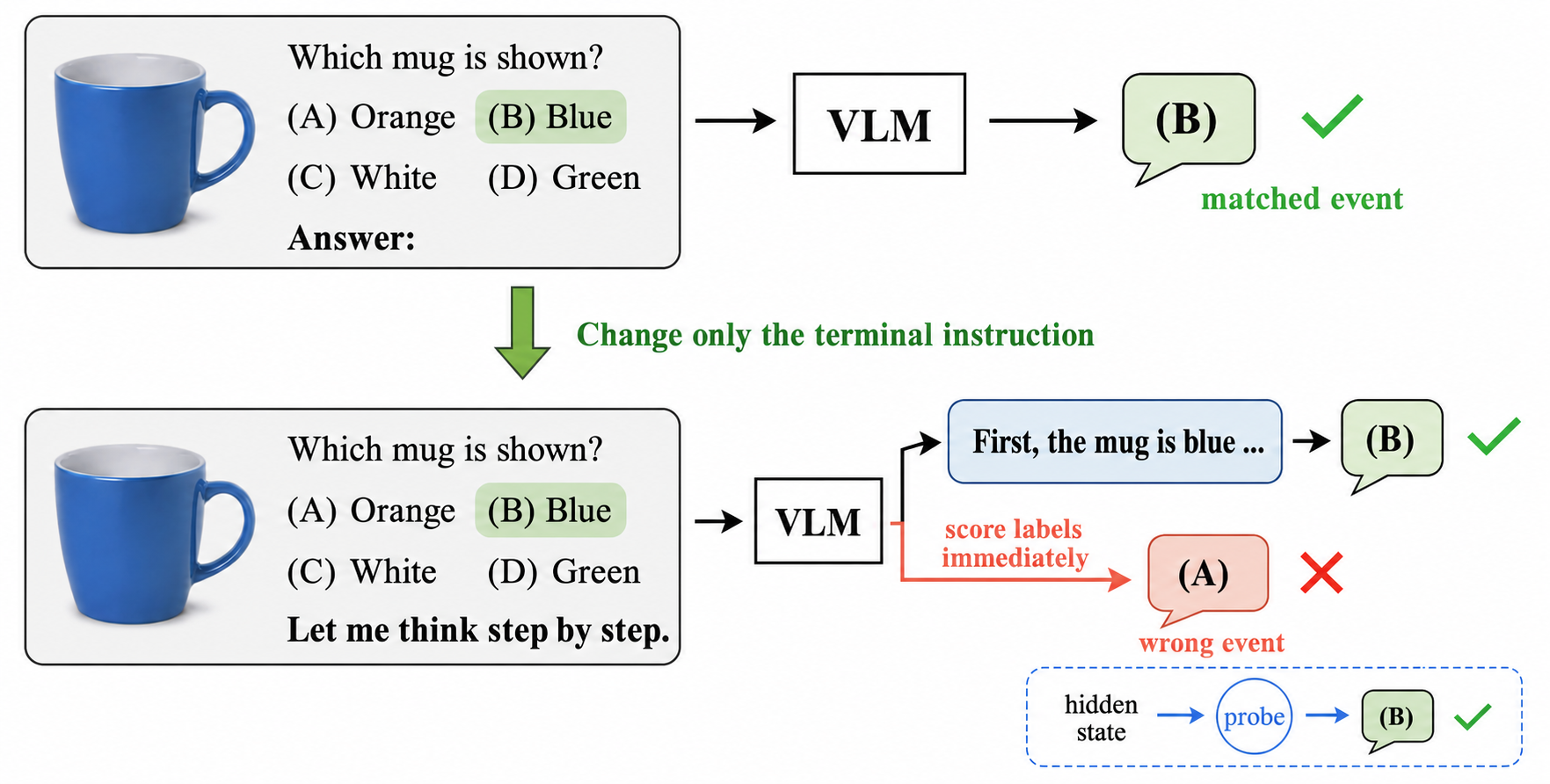}
\caption{A single suffix creates an event mismatch. Direct prompting requests and scores B; the CoT prefix requests a rationale, so immediate label scoring can return A even when free generation and a matched probe recover B.}
\label{fig:summary}
\end{figure}

Distinguishing answer loss from readout failure matters because multiple-choice accuracy can shift for reasons unrelated to model knowledge. Prompt format, answer priors, option order, and the surface form being scored all affect the measured result \cite{zhao2021calibrate,holtzman2021surface,zheng2024large}. Prior work has benchmarked LVLM selection bias and logit correction \cite{atabuzzaman2025benchmarking,tian2025identifying}, analyzed position effects \cite{wang2025eliminating,shi2025judging}, and documented prompt-format flaws in multiple-choice VQA \cite{rosenthal2025unexplored}. What these studies do not isolate is the specific event mismatch considered here: a continuation-inducing suffix is appended, but the evaluator immediately restricts scoring to answer labels before any rationale token is generated. We therefore use condition-matched probes trained on the official training split, selected on validation, and evaluated on a locked test set, together with four controls: option-content permutation, free reasoning before answer extraction, alternative scored output events, and replication across datasets and model families.

The results support readout failure rather than answer loss. A matched-capacity probe recovers most of the lost points from the same final hidden state, and letting the model actually generate the requested reasoning restores most of the direct-answer performance. The collapse recurs, sometimes more severely, on other datasets and models, while one counterexample shows that the suffix can help instead. Figure~\ref{fig:summary} summarizes this \emph{CoT-prefix decodability gap}. The important distinction is that a low immediate label score need not imply that the model has lost the answer; the requested continuation can change which event the native readout is prepared to emit.

Our contributions are threefold. First, we identify and name CoT-prefix scoring as a common evaluation shortcut. Second, we show through condition-matched probes and generation controls that the answer survives the prefix and the failure is localized to the immediate readout. Third, we trace the failure to a shift of probability mass onto continuation tokens, and we recommend that evaluations align the requested output event with the scored output event.

\section{Evaluation Interfaces and Diagnostic Protocol}
\label{sec:protocol}

We compare three ways of obtaining an answer to the same multiple-choice item. Every item contains an image when available, a question, and two to five lettered choices. \textbf{Direct} ends the prompt after the choices and scores the native logits of the valid answer tokens. \textbf{CoT-prefix} appends ``Let me think step by step.'' yet still scores the answer tokens at the very next position, without generating any text. \textbf{Free CoT} lets the model produce the requested continuation and extracts an answer only after generation. The first two conditions differ only in the terminal text, so their contrast isolates the suffix. The third also changes the output event, so its contrast with the second isolates generation itself.

Let $P_c(x)$ denote the complete prompt under condition $c\in\{D,C\}$, $h_c(x)=f_\theta(P_c(x))$ the hidden state at the final input position, and $E_v$ the output embedding of vocabulary token $v$. The model's next-token distribution is
\begin{equation}
 p_c(v\mid x)=
 \frac{\exp(E_v^\top h_c(x))}
 {\sum_{u\in\mathcal V}\exp(E_u^\top h_c(x))}.
 \label{eq:next-token}
\end{equation}
Let $\mathcal A(x)$ be the valid answer-label tokens. Immediate scoring and free reasoning query two different conditional events. The first reads the label distribution at the next position,
\begin{equation}
 \hat y_{\mathrm{imm}}^{(c)}
 =\arg\max_{a\in\mathcal A(x)}p_c(a\mid x),
 \label{eq:imm}
\end{equation}
while the second queries the multi-token event of generating a rationale and then an answer,
\begin{equation}
 \hat y_{\mathrm{free}}^{(C)}
 =\operatorname{Extract}\!\left(
 \arg\max_{y_{1},\dots,y_{T}}\sum_{t=1}^{T}
 \log p_\theta(y_t\mid P_C(x),y_{<t})\right).
 \label{eq:free}
\end{equation}
CoT-prefix scoring applies Eq.~\eqref{eq:imm} with $c=C$ even though $P_C$ requests the event in Eq.~\eqref{eq:free}. We study this event mismatch rather than CoT generation itself. Restricted-token scores are a common task readout \cite{petroni2019language}, but here they are read at a position primed for rationale generation.

The probe replaces the label rows of the vocabulary projection with a learned matrix $W_c\in\mathbb R^{5\times3584}$, where the five rows cover the maximum number of displayed choices and invalid rows are masked at scoring time,
\begin{equation}
 \hat y_{\mathrm{probe}}^{(c)}=
 \arg\max_{a\in\mathcal A(x)}(W_ch_c(x))_a.
 \label{eq:probe}
\end{equation}
For each condition, we extract the final-layer state at the last input position and train a separate linear map from its 3,584 dimensions to five choice logits, with invalid choices masked and the VLM frozen. Training uses cross-entropy with AdamW, at most 50 epochs, and retains the best validation checkpoint. We summarize recovery by
\begin{equation}
\begin{aligned}
 \Delta_C&=\operatorname{Acc}(\hat y_{\mathrm{probe}}^{(C)})
 -\operatorname{Acc}(\hat y_{\mathrm{imm}}^{(C)}),\\
 \Delta_D&=\operatorname{Acc}(\hat y_{\mathrm{probe}}^{(D)})
 -\operatorname{Acc}(\hat y_{\mathrm{imm}}^{(D)}),
\end{aligned}
 \label{eq:recovery}
\end{equation}
and define the condition gap as $\Gamma=\Delta_C-\Delta_D$. A large $\Gamma$ indicates condition-specific recovery by equal-capacity readouts. Probes of this kind are a standard test of linear accessibility \cite{alain2016understanding}. A supervised probe measures what can be read out of a state, not what the model itself uses \cite{hewitt2019designing,belinkov2022probing}, and a large $\Gamma$ does not imply that the two hidden states are identical.

We use the public Qwen2.5-VL-7B-Instruct checkpoint in evaluation mode and all official ScienceQA test questions, including image-present and text-only items. Qwen and LLaVA represent the LVLM family that connects pretrained visual representations with autoregressive language models \cite{bai2025qwen25vl,liu2023llava}. Inputs share the chat template, preprocessing, question, hint, and choices, and only the terminal instruction differs. Each probe has roughly eighteen thousand parameters, and both conditions use the same capacity and selection rule.

We compute native accuracy as a micro-average over items with invalid answer labels masked. Free-CoT extraction happens after generation, and outputs without a recoverable choice count as errors. Each item keeps its natural number of choices, and we also include a full-string control that scores option text instead of short labels. The evaluation is locked as follows. Probes are fit on the image-present portion of the official ScienceQA training split, checkpoints are chosen on the corresponding validation portion, and all reported numbers come from the complete official test split, including text-only items. The test split is never used to fit or select the primary decoder.

To test whether the failure survives changes in answer content, we permute the option contents with five pre-fixed seeds and remap the gold answer. Because letter labels are regenerated in display order, the first slot always remains A. For seed $s$, let $y_i^{(s)}$ be the remapped gold label. We report
\begin{equation}
 \begin{aligned}
 A_c^{(s)}&=\frac{1}{N}\sum_{i=1}^{N}
 \mathbf 1[\hat y_{i,c}^{(s)}=y_i^{(s)}],\\
 B_c^{(s)}&=\frac{1}{N}\sum_{i=1}^{N}
 \mathbf 1[\hat y_{i,c}^{(s)}=\mathrm A],
 \end{aligned}
 \label{eq:shuffle}
\end{equation}
where $B_c^{(s)}$ is the coupled A/first-slot rate. A high $B_C^{(s)}$ after content shuffling indicates a stable default toward this coupled label--position event. This control alone cannot separate position bias from label-token bias. Separating them would require independently counterbalanced labels.

\section{Results}
\label{sec:results}

We first report native accuracy under the two scoring conditions. CoT-prefix scoring reduces native accuracy by more than thirty points, from 80.76\% to 45.48\%, while condition-matched probes recover nearly all of the lost accuracy from the same hidden states. The probe reaches 84.27\% for Direct and 78.94\% for CoT-prefix, so the native/probe gaps are 3.51 and 33.46 points respectively. Across five option-content permutations, Direct accuracy stays at $80.26\pm0.13$\% and CoT-prefix accuracy at $45.78\pm0.66$\%, with A/first-slot rates of $51.71\pm1.06$\% and $93.54\pm0.43$\%. Under uniform permutation the remapped gold label falls in the first slot on 40.0\% of items, since 52.5\% of the split has two choices. Shuffled CoT-prefix accuracy therefore exceeds the pure-first-slot base rate by only 5.8 points. The default is induced by the condition, and it does not depend on the option contents.

Three controls separate the interface mismatch from a genuine loss of reasoning ability, and the same pattern holds on the 2,017-example image-present subset. Letting the model generate the requested chain and then parsing its final choice raises CoT accuracy to 75.24\% on all 4,241 items, with 215 parse failures counted as errors, while the A/first-slot rate falls from 94.08\% to 42.09\%. On the image-present subset, Direct scoring reaches 83.49\%, CoT-prefix falls to 45.27\%, and generation restores 80.61\%, recovering 35.34 of the 38.22 lost points despite 121 unparseable outputs. Forced one-token generation closely reproduces the native accuracies, showing that the failure is tied to the next-token event rather than the generation API. Conversely, scoring length-normalized likelihoods of the full option strings reverses the ordering, 52.32\% for Direct and 55.41\% for CoT-prefix. Recovery also does not depend on one exact suffix: across eleven endings, generated-answer accuracy ranges from 76.75\% to 87.46\%, while answer-eliciting endings such as ``Answer:'' perform well under immediate scoring and continuation-inducing phrases perform substantially worse. The collapse is therefore specific to scoring short answer tokens immediately after a continuation request.

To test whether the failure is localized to the readout, we retrain only the readout. Rank-32 LoRA adapters \cite{hu2022lora} trained on the vocabulary projection alone, using 3,000 examples, frozen transformer layers, ten epochs, and no validation selection, raise CoT-prefix accuracy from 45.48\% to 70.90\%, while Direct stays flat at 80.76\% to 80.60\%. This confirms that the failure is localized to the readout. Table~\ref{tab:scope} repeats the native comparison on further datasets and architectures, with no fitted readouts. These include AI2D \cite{kembhavi2016diagram}, a ten-subject MMMU subset \cite{yue2024mmmu}, MMBench \cite{liu2024mmbench}, both Qwen sizes, and LLaVA-1.5-7B. The collapse is severe but not universal. It reaches 54.02 points on AI2D, yet reverses on MMBench. On AI2D the prefix raises the A/first-slot rate from 24.19\% to 94.66\% for Q7 and from 24.61\% to 66.71\% for Q3. Among examples that flip from Direct-correct to CoT-prefix-wrong, 99.17\% for Q7 and 86.43\% for Q3 land on A.

\begin{table}[t]
\centering
\caption{Cross-dataset/model checks. Q7/Q3 denote Qwen2.5-VL-7B/3B, L7 denotes LLaVA-1.5-7B, and SQA* is the image-present subset. A/first is the CoT-prefix first-slot rate and is omitted when fixed letter labels are unavailable.}
\label{tab:scope}
\begingroup
\ninept
\renewcommand{\arraystretch}{0.90}
\setlength{\tabcolsep}{0pt}
\begin{tabular*}{\columnwidth}{@{\extracolsep{\fill}}llrrrrr@{}}
\toprule
Data & Model & $N$ & Direct & Prefix & $\Delta$acc & A/first \\
\midrule
ScienceQA & Q7 & 4,241 & 80.76 & 45.48 & $-35.28$ & 94.08 \\
AI2D & Q7 & 3,088 & 82.48 & 28.47 & $-54.02$ & 94.66 \\
AI2D & Q3 & 3,088 & 79.02 & 46.73 & $-32.29$ & 66.71 \\
MMMU-10 & Q7 & 286 & 47.90 & 31.82 & $-16.08$ & -- \\
MMBench & Q7 & 4,377 & 93.31 & 99.11 & $+5.80$ & -- \\
SQA* & Q3 & 2,017 & 80.07 & 58.11 & $-21.96$ & 65.10 \\
SQA* & L7 & 2,017 & 66.14 & 63.06 & $-3.07$ & 46.80 \\
\bottomrule
\end{tabular*}
\endgroup
\end{table}

A further boundary condition comes from the number of displayed choices for Q3 on the ScienceQA image subset. For two-choice items, Direct and CoT-prefix accuracy is 85.1\% and 58.2\%, and CoT-prefix selects A 86.3\% of the time. For three choices, the values are 76.7\% and 48.3\%, with 69.9\% A selection. For four choices, the values are 80.6\% and 66.4\%, with 42.1\% A selection. The coupled default weakens as more choices are displayed, and the accuracy gap narrows from 26.9 to 14.2 points. The five-choice slice contains only 38 items, too few to interpret. Taken together, these results show that CoT-prefix scoring systematically diverts the immediate readout toward a coupled default, and that the effect is strongest when few choices are displayed.

\section{Mechanistic and Exploratory Diagnostics}
\label{sec:diagnostics}

\begin{figure}[t]
\centering
\includegraphics[width=0.99\columnwidth]{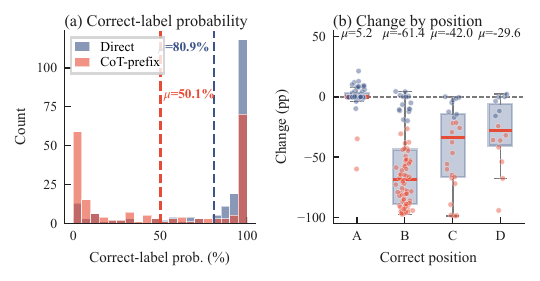}
\caption{Probability diagnostics on 200 fixed items. CoT-prefix preserves correct-label probability in the first slot but suppresses later positions.}
\label{fig:probability}
\end{figure}

Having established that the failure is localized to the readout, we now examine what the prefix changes inside the model. The question is not only whether the answer is still present, but where it remains accessible and how the prefix redirects the readout away from it. On a fixed 200-item subset (Fig.~\ref{fig:probability}), we first measure the position-conditioned probability shift
\begin{equation}
 \begin{aligned}
 I_j&=\{i:r_i(y_i)=j\},\\
 \delta_j&=\frac{1}{|I_j|}\sum_{i\in I_j}
 \left[p_C(y_i\mid x_i)-p_D(y_i\mid x_i)\right],
 \end{aligned}
 \label{eq:position-shift}
\end{equation}
where $j$ indexes the displayed position of the gold label, $p_D$ and $p_C$ denote the probability assigned to the correct label under Direct and CoT-prefix scoring, and the average is taken over items whose correct label appears at position $j$. The statistic therefore isolates how much the suffix changes the correct-label probability at each displayed position, without conflating items with different label placements.

Direct scores concentrate near the correct label, whereas CoT-prefix creates a second mode near zero. The measured $\delta_1$ is near zero, while $\delta_j$ is strongly negative for later positions, which means the suffix suppresses the correct label almost exclusively when that label is not in the first slot. This matches the content-shuffle result above and explains why the A/first-slot rate rises so sharply under CoT-prefix. A representative item illustrates the effect. Direct scoring assigns 0.904 probability to the correct option C, whereas CoT-prefix assigns 0.980 to the A/first-slot option. Free reasoning and the condition-matched probe both recover C, and the probe assigns it probability 0.9996. The requested reasoning succeeds even though the immediate label event fails, so the loss is confined to the readout rather than the representation.

Table~\ref{tab:triggers} tests continuation strength directly on the full test split. Four explicit step-by-step requests cause large drops and A/first-slot concentration, while the minimal cue ``Reasoning:'' remains close to Direct. The effect therefore tracks continuation strength rather than one exact string. A phrase that clearly invites a long continuation suppresses the label readout, whereas a short cue that does not commit the model to generate much leaves the label distribution largely intact.

\begin{figure}[t]
\centering
\includegraphics[width=0.99\columnwidth]{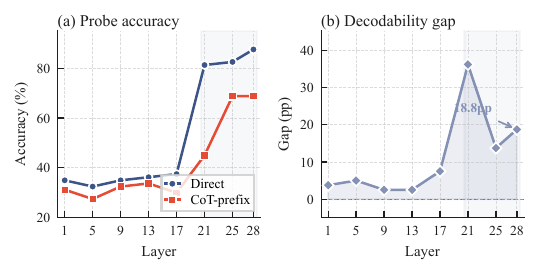}
\caption{Layer-wise diagnostics on 400 fixed items. Probe accuracy rises late, where the Direct--CoT-prefix gap also widens.}
\label{fig:layers}
\end{figure}

\begin{table}[t]
\centering
\caption{Trigger comparison on the ScienceQA test split. $\Delta$acc is relative to Direct; probe columns report condition-matched accuracy and A/first rates (\%).}
\label{tab:triggers}
\begingroup
\ninept
\renewcommand{\arraystretch}{0.88}
\setlength{\tabcolsep}{0pt}
\begin{tabular*}{\columnwidth}{@{\extracolsep{\fill}}lrrrrr@{}}
\toprule
Instruction & \shortstack{Native\\Acc.} & $\Delta$acc & \shortstack{Native\\A/first} & \shortstack{Probe\\Acc.} & \shortstack{Probe\\A/first} \\
\midrule
None (Direct) & 80.76 & $+0.00$ & 52.30 & 84.27 & 35.60 \\
Let me think step by step. & 45.56 & $-35.20$ & 94.08 & 78.94 & 35.96 \\
Let's think step by step. & 49.19 & $-31.57$ & 90.10 & 79.86 & 35.46 \\
Let's solve this step by step. & 52.49 & $-28.27$ & 86.49 & 79.96 & 35.32 \\
Let's reason step by step. & 52.77 & $-27.99$ & 85.76 & 79.89 & 36.95 \\
Reasoning: & 77.18 & $-3.58$ & 56.17 & 82.60 & 40.30 \\
\bottomrule
\end{tabular*}
\endgroup
\end{table}

Let $\mathcal R$ be the fixed set of common reasoning-continuation tokens, such as ``First'', ``Let'', and ``The''. The unrestricted-vocabulary masses, not renormalized over the valid labels, are
\begin{equation}
 M_{\mathcal A}^{(c)}(x)=\sum_{a\in\mathcal A(x)}p_c(a\mid x),
 \qquad
 M_{\mathcal R}^{(c)}(x)=\sum_{v\in\mathcal R}p_c(v\mid x).
 \label{eq:token-mass}
\end{equation}
Table~\ref{tab:competition} separates two effects that would otherwise be conflated. The first is expected once the requested output event changes. Probability mass moves from answer labels to reasoning-continuation tokens: $M_{\mathcal A}$ falls by about fourteen-fold, $M_{\mathcal R}$ rises four-fold, and the mean rank of the correct answer token worsens by more than twenty thousand positions. This shift alone is not a failure; a model prompted to explain should prefer prose tokens over a bare label. The second effect is the failure studied here. Even after restricting attention to the valid label set, predictions concentrate on the coupled A/first-slot label, and correct-label probability is preserved mainly when the gold option occupies that slot (Fig.~\ref{fig:probability}). Thus the suffix does more than move probability mass out of the answer vocabulary. It also changes the conditional distribution within the answer labels, which explains why simply renormalizing the valid choices does not recover the original decision.

\begin{table}[t]
\centering
\caption{Full-vocabulary competition on the ScienceQA image subset ($N=2{,}017$); Change compares CoT-prefix with Direct.}
\label{tab:competition}
\begingroup
\ninept
\renewcommand{\arraystretch}{0.90}
\setlength{\tabcolsep}{2pt}
\begin{tabular}{lrrr}
\toprule
Diagnostic & Direct & CoT-prefix & Change \\
\midrule
Answer mass & $2.15{\times}10^{-8}$ & $1.51{\times}10^{-9}$ & $\downarrow 14\times$ \\
Reasoning mass & $3.73{\times}10^{-2}$ & $1.49{\times}10^{-1}$ & $4.00\times$ \\
Answer rank (mean) & 5,251 & 26,562 & $+21{,}311$ \\
Reasoning top token (\%) & 4.3 & 9.9 & $+5.6$ pp \\
\bottomrule
\end{tabular}
\endgroup
\end{table}

To locate where the answer becomes decodable, we fit the same linear readout to the layer-$\ell$ states $h_c^{(\ell)}$ and compute
\begin{equation}
 \begin{aligned}
 \hat y_{i,c}^{(\ell)}
 &=\arg\max_{a\in\mathcal A(x_i)}
 (W_c^{(\ell)}h_{i,c}^{(\ell)})_a,\\
 A_c^{(\ell)}
 &=\frac{1}{N_{\mathrm{te}}}\sum_i
 \mathbf 1[\hat y_{i,c}^{(\ell)}=y_i],\\
 G^{(\ell)}&=A_D^{(\ell)}-A_C^{(\ell)}.
 \end{aligned}
 \label{eq:layer-gap}
\end{equation}
In the fixed 400-item run (Fig.~\ref{fig:layers}), both $A_D^{(\ell)}$ and $A_C^{(\ell)}$ remain near chance through early and middle layers, then rise sharply. The gap $G^{(\ell)}$ widens only in the late layers, where answer information becomes linearly organized. This layer profile is consistent with the view that answer identity is computed late in the network and that the readout depends on a late-layer representation. The prefix does not remove this representation, but it changes how the final position aggregates it.

A shuffle control confirms that the probes read content rather than label frequencies. Permuting hidden states across examples drops linear accuracy from 86.25\% to 33.75\%, near the empirical random-choice baseline of 32.65\%, and drops the full decoder from 86.25\% to 38.75\%. Probe recovery therefore depends on example-specific state information. Taken together, these diagnostics show that the answer remains linearly organized in the late layers, while the prefix redirects the immediate readout toward continuation tokens. The information is still present, but the scoring interface no longer exposes it. This is why a probe or a generation step can recover the answer while the native label logits cannot.

\section{Conclusion}
\label{sec:conclusion}

We identify CoT-prefix scoring, where an evaluator appends a reasoning instruction but reads answer-label logits before any rationale is generated. On ScienceQA, this convention lowers Qwen2.5-VL-7B accuracy by more than 35 points and drives predictions toward the first slot regardless of its content. Condition-matched probes recover most of the answer from the same hidden states, while free generation restores most of the direct-answer performance. Vocabulary, probability, and layer-wise diagnostics locate the mismatch: the prefix redirects probability mass toward continuation tokens, yet answer identity remains linearly organized in late layers. The failure therefore lies in the alignment between the requested continuation and the immediate scoring event, not simply in whether the representation contains the answer. Its severity varies across datasets and models and is not universal.

Practically, our results suggest a simple evaluation checklist. If a prompt asks for a rationale, the scoring rule should read the token event that corresponds to that request (e.g., the generated answer after the rationale, or a dedicated answer field), rather than the immediate label logits at the first post-prefix position. When label-logit scoring is required for compatibility, the prompt should avoid continuation-seeking instructions, or explicitly constrain the next token to be an answer label. Reporting should also specify the exact interface (labels vs.
free-form generation), the label tokenization, and whether decoding is conditioned on an intermediate rationale, since these choices can dominate measured accuracy.

Our study is limited to a set of multiple-choice benchmarks and two VLM families. Linear probes establish decodability rather than causal use by the native decoder, and content shuffling cannot fully separate position bias from label-token bias without independently counterbalanced labels. We also study only a limited set of reasoning instructions and answer interfaces. These limits do not change the main evaluation lesson: when a prompt requests a rationale, the scorer should evaluate the event the model is actually asked to produce. Future work should test broader model families, open-ended tasks, and training or decoding interventions (e.g., interface-aware instruction tuning or constrained decoding) that retain the benefits of reasoning prompts without breaking immediate answer readout. More broadly, evaluation protocols should treat the prompt-and-scorer pair as part of the method specification, not an interchangeable implementation detail.

% Keep all technical content within the first four pages and start references on page 5.
% If the last technical column has a small amount of remaining space, \vfill absorbs it
% without changing the font size or squeezing the main text.
\vfill
\pagebreak

\bibliographystyle{IEEEbib}
\bibliography{refs}
\end{document}